\documentclass[journal,onecolumn]{IEEEtran}

\ifCLASSINFOpdf

\else

\fi

\usepackage{graphicx}
\usepackage{adjustbox}
\usepackage{amsmath}
\usepackage{multirow}
\begin{document}
\graphicspath{{figures/}}

\title{Language Identification with \\
Deep Bottleneck Features}

\author{Zhanyu~Ma, Hong~Yu
\thanks{Z. Ma,  and H. Yu are with Pattern Recognition and Intelligent System Lab., Beijing University of Posts and Telecommunications, Beijing, China.}
}

\maketitle

\begin{abstract}
In this paper we proposed an end-to-end short utterances speech language identification(SLD) approach
based on a Long Short Term Memory (LSTM)  neural network which is special suitable for SLD  application in intelligent vehicles.
Features used for LSTM learning are generated by a transfer learning method. Bottle-neck features of a deep neural network (DNN) which are trained for mandarin
acoustic-phonetic classification are used for LSTM training. In order to improve the SLD accuracy of short utterances a phase vocoder based time-scale modification(TSM)
method is used to reduce and increase speech rated of the test utterance.
By splicing the normal, speech rate reduced and increased utterances, we can extend  length of test utterances so as to improved improved the performance of
the SLD system. The experimental results on AP17-OLR database shows that the proposed methods can improve the performance of SLD, especially on short utterance with 1s and 3s durations.
\end{abstract}


\IEEEpeerreviewmaketitle

\section{Introduction}
\IEEEPARstart{T}he task of speech language identification (SLD) is to automatically  recognize the language of the given spoken utterance which has been widely used as the front-end of the mixed lingual speech recognition(SR) system~\cite{zissman1996comparison}.
The language of the speech utterance should be recognized firstly,and then the SR system can call the corresponding decode to translate the input speech utterance into right text.

There are thousands of languages in the world and each language has different distinguishing features, many different researches have been worked on developing a universal, quick responsive and effective SLD system\cite{tang2018phonetic}. Generally speaking, researches about SLD focuses on two domain, in the front-end domain, researches want to find features which can express the difference
between different languages and in the back-end domain, effective classification schemes are needed to distinguish diverse languages.

In the feature domain, raw acoustic features. e.g., linear predictive coding (LPC), filter bank feature, and formation features are early considered~\cite{cimarusti1982development}\cite{foil1986language}.
Then, the performance of dynamic features which include temporal information are also investigated\cite{torres2002approaches}.
Prosody information, such as the patterns of duration, pitch and stress of languages, is usually used as additional knowledge to improved the performance of raw acoustic features~\cite{muthusamy1993segmental}\cite{hazen1997segment}\cite{rouas2003modeling}.
Token based features, such as phone, syllables and words sequences which contain hight-level character information, are also used to realize the SLD function~\cite{zhu2005different}
\cite{schultz1996lvcsr}\cite{hieronymus1997robust}.

In the classifier domain, when using acoustic or prosody features as front-ends, strong statistical models are usually selected to build the SLD system.
In paper~\cite{torres2002approaches}\cite{willmore2000comparing}, different languages are modeled by Gaussian mixture models(GMMs) and log-likelihood ratios are used to make languages
identification decision.  Hidden Markov models(HMMs) trained by speaker and text independent acoustic feature sequences of different languages are used to construct SLD system in paper
~\cite{wong2004automatic}\cite{nakagawa1992speaker}.
Neural networks and support vector machines are also used as the back-end to classify different speech languages in\cite{muthusamy1993segmental}\cite{kwasny1992identifying}\cite{campbell2004language}.
The i-vector based method which have been successfully used in speaker verification tasks are also used to express different languages, following with a task-oriented
probabilistic linear discriminant analysis (PLDA) scoring method, the i-vector-PLDA model also achieved significant success in SLD tasks\cite{dehak2011language}\cite{martinez2011language}.  When selecting token based features, e.g., phone sequences,  as front-ends, n-gram language models(LM) are needed as back-end for each target languages to evaluate the confidence that the input speech match that language, which is called  phone recognition and language modelling (PRLM)\cite{tang2018phonetic}.
Different multiple PRLMs based on parallel phone recognition and  phone selection on multilingual phone set were discussed in paper~\cite{zissman1996comparison}\cite{matejka2006brno}\cite{hazen1997segment}.

Recently,  with the developing of deep learning technology, many deep neural networks(DNN) base solution are also involved into SLD takes.
In paper~\cite{lopez2014automatic}, a fully connected feed-forward neural network trained by $21$ frames  stacking perceptual linear prediction(PLP) features are used to classify languages directly. In~\cite{lozano2015end}\cite{jin2016lid}, convolutional neural networks trained by Mel frequency cepstral coefficient(MFCC) and PLP feature maps are applied to build language classifiers. In~\cite{matejka2014neural}\cite{song2013vector}, a DNN is trained to generate frame-level bottleneck (BN) features and these features are used to train an i-Vector based SLD systems. Segment-level X-vectors which are built by mean and standard deviation of BN features are also used in language identification~\cite{snyder2018spoken}.
Recurrent neural networks (RNN) which can model the temporal information of features are also widely used in SLD takes.
In~\cite{gonzalez2014automatic} and ~\cite{fernando2017bidirectional}, long Short-Term Memory(LSTM) and bidirectional LSTM (BLSTM) neural networks are trained to recognize different languages.
Many published results show that the DNN based SLD methods perform better than statistic models based methods, such as GMMs and i-vector,especially on short utterances which can not
supply sufficient statistical information. While when the length of input utterances shorter than three seconds, even the performance of DNN based SLD systems will decline sharply.
In the verbal system of intelligent vehicles, most of the communications are short utterances, so it is very important to build a SLD system that is suitable for very short utterances.
The mainly problem of SLD on short utterance is the inadequate information of input speeches, in order to solve this shortcoming, we build a end-to-end SLD system based on transfer learning features and time-scale modification(TSM).

We use a TSM method to expend the length of short input utterances.
The speech rate of test short utterances are adjusted by phase vocoder method, by splicing the  original speech with a speech rate increased speech and a speech rate decrease speech,
the lack of information of short utterance can made up.
PLP features concatenating with pitch features generated by length expended speeches are used to train the language classifiers.
Many researches shows that the DNN-BN feature generated by a phonetic classifier have more information than raw acoustic features,
so we use PLP+pitch features to train an mandarin phoneme classifier firstly and the pre trained DNN is used to extract DNN-BN features which including more useful information.
In order to suite for short duration speeches, the generated DNN-BN features frames are packed into small blocks with 100 frames. In order to fit short input utterance with frame length less than 100, we use repeatedly padding method to fill the gaps.
Feature blocks are feeded into two layer LSTMs, which are suitable for model feature sequences.
Because the output of the last frame contents information of the whole block, a softmax layer is connected with the last frame of LSTM outputs to realize the language classification task.

In the following sections, we first introduce the TSM method used for short utterance length extending in Section~\ref{TSM}.
The strutters of  neural networks used for DNN-BN features extracting and language classification are described in Section~\ref{NN}.
In Section~\ref{experiment}, we introduce the experimental configuration including the database used for evaluating SLD models and parameters of SLD models.
In this section, we make a comparison between the proposed TSM-DNN-BN-LSTM SLD model and two baseline systems and analyze the experimental results.
Some conclusions are made in section~\ref{conclusion}.

\section{Time-scale Modification}
\label{TSM}
Many published results show that the accuracy of SLD systems will decreased heavily by the shorten of input utterances. In order to solve this problem,
in this paper, we use a  time scale modification (TSM) technology to adjust the speech rate of the input  utterance. By concatenating speeches with different speech rates, we can extend
length of input utterances and increase  information content of  test speeches, so as to improved the performance of the SLD system.

TSM technology can change speech rates by changing the length of speeches. In this section we will introduce a classical TSM method, phase vocoder, which can modify the speech rate of
input speech without badly damage on pitch and prosody information.

Speech rate usually refers to the speed of pronunciation.  Irregular speech rates will decrease the accuracy of continuous SR systems~\cite{yuan2006towards}\cite{goldwater2010words}.
In speaker verification (SV) system, the mismatch of speech rates between enrollment and verification speeches will also degrade the performance SV system~\cite{van2007speech}.
During acoustic features extraction, the speed rate information will be contained into extracted features, inevitably.
In SR and SV tasks, the mismatch of speech rates between training and testing data will decrease the performance and we need to  restrain these mismatches~\cite{wang2007robust}\cite{nejime1996portable}.
While,in the SLD task, abundance combinations of speech rates information will improve the performance SLD system\cite{xiaoxiao2018expanding}.

%

The TSM method can modify the speech rate without changing spectral information, e.g., fundamental frequency and formant. Recently, many different TSM has been proposed and in this paper
we select the phase vocoder method. We can use three steps to modify speech rate.
\begin{figure}
\center
\includegraphics[width=0.30\textwidth]{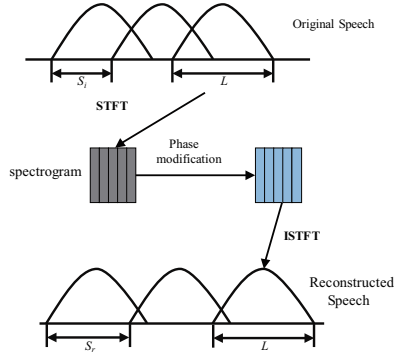}
\caption{The processing flow of phase vocoder.}
\label{fig:1}
\end{figure}
Firstly, the input utterance was segmented into frames with duration,  $L$  and step size, $S_i$. For each frame, a Hanning window was used to reduce the high frequency components.
Short-time Fourier transform (STFT) is applied on each frame and the time-frequency information $X(\lambda ,k)$can be generated by equation~\ref{FV_1}.
\begin{equation}
\label{FV_1}
X(\lambda ,k)=\sum_{n=0}^{L-1}x(\lambda S_i+n)h(n)e^{-j(2\pi kn/N)},
\end{equation}
where x stands for the input speech, $h$ is the window function,  $\lambda$ is the frame index, $k$ means the frequency bin index and N is the point number of discrete Fourier transform.

Secondly, compute the amplitude $|X(\lambda ,k)|$ and phase $\theta ({\lambda ,k})$ of $X(\lambda ,k)$ and then use the idea introduced
in paper~\cite{Laroche1999Improved} to modify the phase into $\theta ^{'} ({\lambda ,k})$.

Finally, inverse short-time Fourier transform (ISTFT) is used to reconstruct time-domain frame $y(\lambda)$ with new phases.
\begin{equation}
\label{FV_2}
y(\lambda )= ISTFT(|X(\lambda ,k)|e^{j\theta ' ({\lambda ,k})}).
\end{equation}
As shown in Fig.~\ref{fig:1} ,by summing reconstructed frames using $S_r$ as step size, we can modify the length of input speeches.

When the step size, $S_r$, in the reconstruction processing is shorter than the step size, $S_i$ in frame segmenting procedure, speech rates of new generated speeches is increased.
On the contrary, we can reduce the speech rate of the input speech. We can define the changing rate of
original speech rate as
\begin{equation}
\label{FV_3}
\alpha = \frac{S_i}{S_r}.
\end{equation}
and the length of the reconstructed speech, $\widetilde{y}$, is:
\begin{equation}
\label{FV_4}
length(\widetilde{y}) = \frac{length(x)}{\alpha }.
\end{equation}
\begin{figure}
\center
\includegraphics[width=0.48\textwidth]{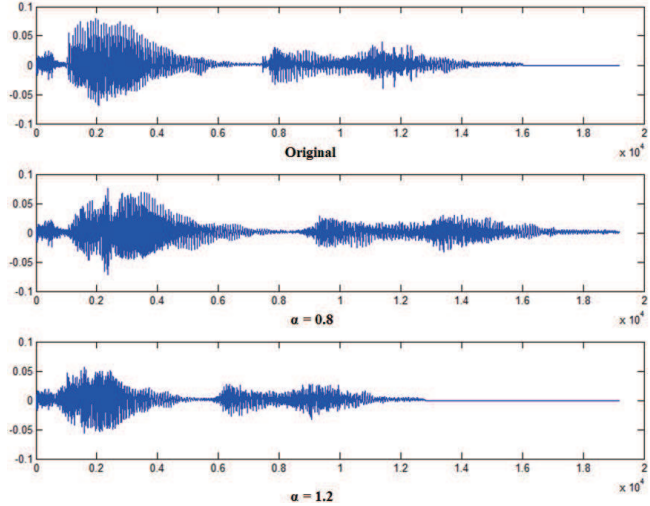}
\caption{Waveform of original, speech rate decreased ($\alpha = 0.8$) and speech rate increased signals ($\alpha = 1.2$).}
\label{fig:2}
\end{figure}
\begin{figure}
\center
\includegraphics[width=0.42\textwidth]{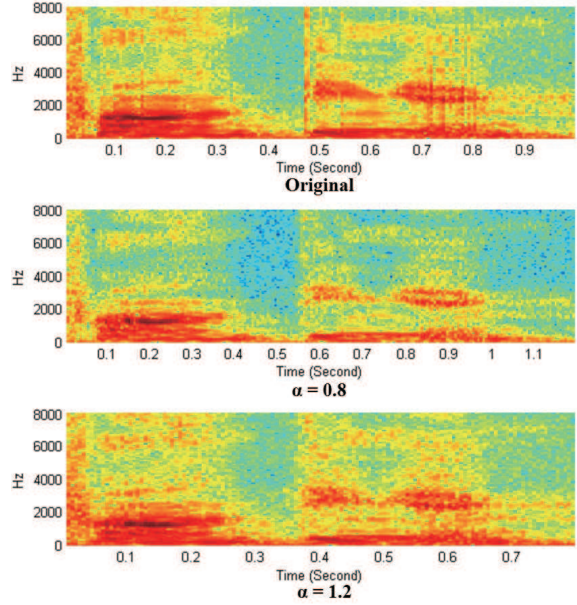}
\caption{Spectrogram of original, speech rate decreased ($\alpha = 0.8$) and speech rate increased signals ($\alpha = 1.2$).}
\label{fig:3}
\end{figure}

Frame duration $L$ is set as 2048 (0.128s) and  discrete Fourier transform number N is also set as 2048.  The step size of reconstructed speech, $S_r$ is set as 512 (32ms).
We can observe that when ignoring frame numbers difference before and after TSM and
aligning three spectrograms to the same size, three aligned spectrograms are very similar.
It means that using TSM method to extending or shorting the length of the same speech will not affect the frequency domain information of original signals,obviously.

Speech rated changed speeches generated by TSM method have less  distortion, by concatenating these speeches with original speech can supply more useful information which is helpful to
SLD tasks.

\section{Structure of Neural Network based SLD model}
\label{NN}
In order to make the proposed SLD model suitable for short utterances, in front-end we use TSM method to extend the length of input signals, so as to increase information helpful to language recognition and in back-end we design a DNN based module to generate more meaningful feature to improve the accuracy of SLD.

\subsection{DNN-BN feature extractor}
In the early stage of SLD realization, researches tend to train statistical models to present different languages.
Specially, i-vector based models trained only by raw  acoustic features archived good performance.
The performance of statistical models is closely related to the frame length of evaluate utterances. In short duration situation, limited testing feature frames
can not supply enough statistical information to SLD models, which will  severely affecting the accuracy of language recognitions.
In order to make the trained model can adapt to short utterances, we need to adopt some other features including more language discriminative information rather than raw acoustic features.

The theoretical foundation of PRLM model is that languages are discriminated by  phonetic properties, this encourages us to build a feature extractor that can extract phonetic information.
We build a DNN using morpheme labels as training target to extract DNN-BN features.
In order to consider the temporal information, we use $M$ concatenated acoustic feature frames as input. The DNN include $L$ hidden layers, the activation function of the bottom $L-1$ layers is sigmoid function and the top hidden layer is a linear layer which is connected with a softmax output layer.

Because outputs of the top hidden layer is nearest to morpheme classification outputs and include abundant morpheme discriminative information, we use these outputs as DNN-BN features to train language  classifiers.
Comparing with unit-level token features which are also include phonetic information, frame-level DNN-BN features have higher temporal resolutions and are more suitable for short utterance SLD.

In some paper, researcher tend to use language label as trained target to train DNN classifiers~~\cite{lopez2014automatic} and want to use a DNN to learn language discriminative information directly, while the language label is too coarse to supply enough supervisory information.
Morpheme labels can provide strong supervision and lead the DNN to learn more useful information layer by layer. Morpheme information also  has close
correlation with language recognition task. Using morpheme label to generate feature for language identification tasks can be thought as some kinds of transfer learning where a related
task is used to pre train a model for another tasks.

DNN-BN feature extractor supervised by morpheme labels also has the benefit of cross language, which means that the DNN trained by one language can learn features to recognize other language. It is important for uncommon language recognition which do not have enough training data.
\subsection{Feature block}
\label{block}
In order to make the designed SLD model suitable for short utterances, we package DNN-BN features into short-time blocks.
In testing phase, input utterances are recognized as block-level and the average log-likelihood value of all blocks are used as speech identification score.

A DNN-BN feature sequence is segmented in to blocks with block size $L_b$ and step size $S_b$.
For long utterances with framed number bigger than $L_b$, the last $L_b$ frames are  packaged separately as a new block.
For short utterances with framed number less than $L_b$, repeating method is used to increase the frame number of input features and then the extended features are
packaged following the method used for partitioning long utterances.

In training step,packaging extracted features into partial overlapping small blocks can make the utmost of limited training data and make the trained model adapt to short utterance.
When testing short utterances, the features repeating method can supply more effective information to trained language classifiers.

\subsection{LSTM-based Language Classifier}
\label{lstms}

The LSTM is a special kind of RNN, which can learn long-term dependencies.
The memory blocks and gates in LSTM cells make it can avoid the long-term dependency problem.
LSTMs are suitable for modeling feature sequences and have good performance on SLD tasks\cite{tian2016investigation}\cite{gonzalez2014automatic}.
In this paper we train a LSTM model with two layers to realize the language classification.
As shown in Fig.~\ref{fig:fig_lstm}, the language classifier is trained by DNN-BN feature blocks with $L_b$ frames.
The LSTM based model can build a mapping from input feature sequences $\left ( x_1,..x_{L_b} \right )$ to hidden layer outputs $\left ( h_1,..h_{L_b} \right )$.
Because the last output frame of the top hidden layer, $h_{L_b}$, is generated by all input features, it can
present the information of the whole input feature block.
We feed $h_{L_b}$ to a full connect layer with  rectified linear units (Relu) as activation function and use a softmax layer to classify different languages.

Insight into the LSTM cell, a popular "peephole connections"~\cite{gers2000recurrent} structure is selected.
As shown in Fig.~\ref{fig:fig_lstm},  square icons stand for neural network layers and circular icons mean point-wise operation.
The associated computation is given as fellows:
\begin{equation}
\label{lstm_1}
f_t = sigmoid\left ( W_f \cdot [c_{t-1},h_{t_1},x_t] + b_f\right ),
\end{equation}
\begin{equation}
\label{lstm_2}
i_t = sigmoid\left ( W_i \cdot [c_{t-1},h_{t_1},x_t] + b_i\right ),
\end{equation}
\begin{equation}
\label{lstm_3}
\tilde{c}_t = tanh\left ( W_c\cdot [h_{t-1},x_t]+b_c \right ),
\end{equation}
\begin{equation}
\label{lstm_5}
c_t = f_t\ast c_{t-1} + i_t\ast \tilde{c}_t,
\end{equation}
\begin{equation}
\label{lstm_6}
o_t = sigmoid\left ( W_o \cdot [c_{t},h_{t-1},x_t] + b_o\right ),
\end{equation}
\begin{equation}
\label{lstm_7}
h_t = tanh(c_t)\ast o_t,
\end{equation}
where "$\cdot$" means matrix multiplication and "$\ast$" means points-wise multiplication.

\section{Experiments}
\label{experiment}
\subsection{Database}
The proposed SLD model is evaluated on the
AP17-OLR databases which are used for the second oriental language recognition challenge~\cite{tang2017ap17}.
The database  is  originally created by Speechocean and  Multilingual Minorlingual Automatic
Speech Recognition (M2ASR). In the databases, there are totally 10 languages including  Kazakh in China (ka-cn) , Tibetan in China (ti-cn),
Uyghur in China (uy-id),
Cantonese in China Mainland and Hongkong (ct-cn),
Mandarin in China (zh-cn),
Indonesian in Indonesia (id-id),
Japanese in Japan (ja-jp),
Russian in Russia (ru-ru),
Korean in Korea (ko-kr),
and Vietnamese in Vietnam (vi-vn).
\begin{table}[!htb]
\renewcommand{\arraystretch}{1}
\caption{Description of experimental database.}
\label{tab:database}
\centerline{
\begin{tabular}{| c | c | c |c|c|}
\hline
& \multicolumn{2}{|c|}{train/dev}& \multicolumn{2}{|c|}{test}\\
\hline
Language.     & Speaker & Total utt. &Speaker &Total utt.\\
\hline
ka-cn  &86 &4200  &86   &1800\\
ti-cn  &34 &11100 &34   &1800\\
uy-id  &353 &5800 &353  &1800\\
ct-cn  &24  &7559  &6   &1800\\
zh-cn  &24 &7198   &6   &1800\\
id-id  &24 &7671  &6    &1800\\
ja-jp  &24 &7662  &6    &1800\\
ru-ru  &24 &7109  &6    &1800\\
ko-kr  &24 &7196  &6    &1800\\
vi-vn  &24 &7200  &6    &1800\\
\hline
\end{tabular}
}
\end{table}

The database is divide into a train/dev part and a test part, details about speaker number and total utterances of each language are described in Table~\ref{tab:database}.
Male and female speakers and utterances of each speaker are balanced. Speakers in train/dev and test subsets have no overlap.

All utterances were recorded by mobile phones, with a sampling rate of 16kHz
and a sample size of 16 bits. In the train/dev subset each language has about 10 hours recordings can be used for SLD model training.
In order to investigate the performance of trained SLD model on short duration signals, the AP17-OLR database also supply some short duration subsets,
including train-1s, train-3s, dev-1s, dev-3s, test-1s and test-3s, which are randomly segmented from train/dev and test subset, respectively.

\subsection{Experimental configurations}

In front-end, acoustic and prosody features are used for SLD model training. The input utterances are segmented into frames with 25ms length and 10ms step size, for each frame
150 dimensional PLP coefficients (50 + $\Delta$  + $\Delta \Delta$ ) concatenating with 3 dimensional pitch features are extracted.
A global mean and  variance vectors are use to normalize extracted features. A pertained DNN base voice activity detector (VAD) is used to remove silence frames.

All the  neural networks are trained by Tensorflow~\cite{abadi2016tensorflow}.
In the DNN-BN feature extractor training phase, as shown in fig.~\ref{fig:fig_DNN}, 11 frames concatenating PLP+pitch feature are used for training data. The
phoneme discrimination DNN include five hidden layers and the nodes number of each hidden layer are all set as 512 which means the dimension of DNN-BN features generated by the linear outputs of the top hidden layer is also 512.
About 500 hours Mandarin Chinese speech collected from Sogou speech input method platform are used to train the phoneme discriminator. Input features are tagged to 6294 triphone label by a trained acoustic model.
The DNN based phoneme classifier is trained by 1683 dimension (153 $\times$ 11 = 1683 ) features and 6294 dimension target labels, using cross entropy as cost function
and  stochastic gradient descent (SGD) as optimization method. The learning rate is set as 0.001,  the training epoch is set as 50, and the min-batch size is set as 256.

After phoneme classifier training, the trained DNN is used as a feature extractor to generate DNN-BN features.
As described in Section~\ref{block}, produced features with 512 dimensions are segmented into blocks with block size $L_b = 100$ (about one second) and step size $S_b = 50$.

As described in Fig.~\ref{fig:fig_lstm}, packaged short-time feature blocks are send into a language classifier with two LSTM layers. The output nodes number of two LSTM
layers are all set as 512. The nodes number of Relu layer is set as 1024. The dimension of softmax  layer is 10 which stands for languages to be identified.
In language classifier, we also select cross entropy as cost function and an Adam optimizer is used to update parameter in language classifier.
The learning rate is set as 0.0002 and the training epoch is set as 50.  During training only the parameters in the language classifier parts are updated the parameters in
DNN-BN feature extractor part are fixed.

\subsection{Baseline systems}
We build two baseline SLD systems base on i-vector model and LSTM model trained by PLP + pitch features with 153  dimension.

In the i-vector model, the universal background model(UBM) with 2048 Gaussian mixtures are trained by utterances in AP17-OLR database.
The dimension of i-vectors is set as 400.
The mean i-vector of one language in the train/dev subset can be to model that language.
The score of a test utterance on a particular language can be
computed by the cosine distance between the i-vector of the test speech and
the  language model i-vector generated from train/dev subset.

The structure of the baseline LSTM model is similar as the language classifier described in Section~\ref{lstms}.
Instead of DNN-BN features, packaged raw PLP + pitch feature blocks with 100 frames length and 50 frames step size are used for model training.
Mean log-likelihood, computed by outputs of the softmax layer is used as language identification scores.

\subsection{Experimental results}
As in LRE15, performances of different SLD systems are evaluated by  $C_{avg}$ and equal error rate (EER).
The pair-wise loss that composes the missing and false alarm probabilities for a
particular target/non-target language pair is defines as:
\begin{equation}
\label{lstm_1}
C(L_t,L_n) = P_{Target}P_{Miss}(L_t) + (1-P_{Target})P_{FA}(Lt,Ln),
\end{equation}
where $L_t$ and $L_n$ are the target and non-target languages,
respectively; $P_{Miss}$ and $P_{FA}$ are the missing and false alarm
probabilities, respectively. $P_{target}$ is the prior probability for
the target language, which is set to 0.5 in the evaluation.
$C_{avg}$ is defined as the average of the above pair-wise performance:
%
\begin{equation}
\begin{aligned}
& C_{avg}= \frac{1}{N} \{  [ P_{Target}\cdot \sum_{L_t}P_{miss}(L_t)   ] \\
         & + \frac{1}{N-1} [ (1-P_{Target})\cdot \sum_{L_T}\sum_{L_N}P_{FA}(L_t,L_n)] \},
\end{aligned}
\label{MAN_CE1}
\end{equation}
where $N$ is the number of languages.

\begin{table}[!htb]
\renewcommand{\arraystretch}{1}
\caption{Comparison of different SLD models on $C_{avg}$ and EER ($\%$).}
\label{tab:result}
\begin{adjustbox}{max width=0.45\textwidth}
\centerline{
\begin{tabular}{| c | c | c |c|c|c|c|}
\hline
\multirow{ 2}{*}{Models} & \multicolumn{2}{|c|}{test-all}& \multicolumn{2}{|c|}{test-3s} & \multicolumn{2}{|c|}{test-1s}\\
\cline{2-7}
            & $C_{avg}$ & EER & $C_{avg}$ & EER & $C_{avg}$ & EER \\
\hline
i-Vector         &0.063   &6.94    &0.075      &8.67    &0.189       &17.24\\
LSTM             &0.092   &9.64   &0.121       &11.05    &0.136      &14.3\\
\hline
DNN-BN-LSTM      &0.012      &1.94    &0.082         &4.23     &0.073      &9.42\\
TSM-DNN-BN-LSTM  & \textbf{0.006}       & \textbf{0.08}   & \textbf{0.053}           & \textbf{2.62}     & \textbf{0.069}   & \textbf{6.76}\\
\hline
\end{tabular}
}
\end{adjustbox}
\end{table}

Performances of different SLD models are evaluated on full time test subset ,test-all and two short duration subset, test-3s and test-1s.
Utterances level $C_{avg}$ and EER of different models are shown in Table~\ref{tab:result}.

Firstly, we compared the performance of two baseline systems, it can be observed that on the full time data sets,
the statistical-based i-vector model performs a little better than neural network based LSTM model, while in the short  duration test-1s,
because the test utterance can not supply enough  statistical information, the LSTM model perform better than the i-vector.

Secondly, we change the training feature from raw acoustic features to DNN-BN features (DNN-BN-LSTM model), the performance of SLD has been significantly improved.
It indicates that phoneme distinguishing features are very useful for the SLD task.
When we have abundant training data, more complex targets labels can help neural networks to learn features with richer information.

Then, as described in Section\ref{TSM}, the phase vocoder based TSM method is used to extend the length of test utterances (TSM-DNN-BN-LSTM model).
Here, we set the speech rate changing parameter $\alpha$ as 0.8 and 1.2, which means the original speech are concatenate with a speed increased and a speed decreased speech.

From the results in Table~\ref{tab:result} we can see, the TSM based length expending method can improve the accuracy of speech identification.
Without changing parameters of trained neural networks, just by simple preprocessing of input waveform signals, the error rate of SLD system can decline about 50 \% on long duration
data set (test-all, test-3s) and about 30 \% on very short duration speeches (test-1s).

In order to investigate, the affect of speech rate to SLD accuracy, we try some different speed rate changing combination and evaluate their performances on test-1s data set.

\begin{table}[!htb]
\renewcommand{\arraystretch}{1}
\caption{Comparison of speed rate changing combination on test-1 data set.}
\label{tab:result_speed}
\centerline{
\begin{tabular}{| c | c | c |c|c|}
\hline
 $(\alpha_1,\alpha_2)$ & (0.8,1.2)& (1.1,1.2) & (0.8,0.9) & (0.7,1.3)\\
\hline
$C_{avg}$              & \textbf{0.069}   &0.075       & 0.082    &0.075 \\
EER($\%$)              &  \textbf{6.76}   & 7.02       & 7.17     & 7.01 \\
\hline
\end{tabular}
}
\end{table}

From the results in Table~\ref{tab:result_speed} it can be find that, concatenating some speech changed utterances together can improve the accuracy of SLD comparing with the original short signals.
Splicing the original speech with a speech rate increased and a speech rate decreased signal can improve the
performance better than splicing two  speech rate increased speech or two  speech rate decreased speech.
The speech rate changing should be moderate, too big speech rate changing will decrease SLD accuracy.

\section{Conclusion}
\label{conclusion}
In this paper we propose a end-to-end an end-to-end speech language identification(SLD) model. Three measures are used to make the trained model can suitable to short utterances.
In the waveform domain, we use a time-scale modification(TSM) method to extend the length of input utterances. In the feature domain, we use the transfer learning idea to train a deep
phoneme classifier, bottleneck features of the phoneme classifier which include phoneme discriminative information are used to train language classifiers.
In the language classifier domain, a LSTM base classifier are trained by short time feature blocks which can  make the trained model fitting for short duration inputs.
The experimental results on AP17-OLR database show that comparing with the i-vector model and simple LSTM
model, the proposed method can significantly enhance the perforce SLD, especially on short duration utterance.
The structure of proposed SLD model is very simple, the trained model only occupy about 20M  hard disk space. The improvement measures on waveform can avoid the changing on SLD model,
the short-time block segmentation idea can improve the operation speed of LSTM based language classifier.
All the things are suitable for SLD tasks in intelligent cars, which need a small model and quick responses.


%

%
%
%
%
%

\ifCLASSOPTIONcaptionsoff
  \newpage
\fi












\end{document}